# A Novel Partially-decoupled Translational Parallel Manipulator with Symbolic Kinematics, Singularity Identification and Workspace Determination

Huiping Shen[1], Yinan Zhao[2], Ju Li*[1], Guanglei Wu[3], Damien Chablat[4]

[1]School of Mechanical Engineering, Changzhou University, Changzhou 213016, China

[2] Department of Mechanical Engineering, Nanjing Vocational Institute of Mechatronic Technology, Nanjing, 211300, China

[3]School of Mechanical Engineering, Dalian University of Technology, Dalian 116024, China

[4]Laboratoire des Sciences du Numérique de Nantes (LS2N),

UMR CNRS 6004, 1 rue de la Noe, 44321 Nantes, France

shp65@126.com, 936515322@qq.com, wangju0209@163.com, gwu@dlut.edu.cn, Damien.Chablat@cnrs.fr

**Abstract**

This paper presents a novel three-degree-of-freedom (3-DOF) translational parallel manipulator (TPM) by using a topological design method of parallel mechanism (PM) based on position and orientation characteristic (POC) equations. The proposed PM is only composed of lower-mobility joints and actuated prismatic joints, together with the investigations on three kinematic issues of importance. The first aspect pertains to geometric modeling of the TPM in connection with its topological characteristics, such as the POC, degree of freedom and coupling degree, from which its symbolic direct kinematic solutions are readily obtained. Moreover, the decoupled properties of input-output motions are directly evaluated without Jacobian analysis. Sequentially, based upon the inverse kinematics, the singular configurations of the TPM are identified, wherein the singular surfaces are visualized by means of a Gröbner based elimination operation. Finally, the workspace of the TPM is evaluated with a geometric approach. This 3-DOF TPM features less joints and links compared with the well-known Delta robot, which reduces the structural complexity. Its symbolic direct kinematics and partially-decoupled property will ease path planning and dynamic analysis. The TPM can be used for manufacturing large work pieces.

**Introduction**

The 3-DOF translational parallel mechanism (TPM) has significant potential in many industrial applications. It can be classified by actuating modes, i.e., linearly actuated by prismatic joints and rotary actuated by revolute joints. A well-known design of 3-DOF TPM is the Delta Robot, which was proposed by Clavel [1]. The Delta-based TPMs have been developed with alternative prismatic actuated joints [2-3] later. Design optimization of TPMs based on the Jacobian matrices have been carried out [Erreur ! Source du renvoi introuvable.-6] for structural design. Tsai et al. [7] proposed a 3-DOF 3-UPU[1] TPM with three identical limbs consisting of universal-prismatic-universal joints in serial. Li et al. [8-9] proposed a new 3-DOF 3-UPU TPM and analyzed its instantaneous motion performance. In [10, 11], the authors proposed a 3-RRC TPM and investigated the kinematics and workspace. Kong et al. [Erreur ! Source du renvoi introuvable.] proposed a 3-CRR mechanism with good motion performance, free of singular postures. Yu et al. [Erreur ! Source du renvoi introuvable.] carried out a comprehensive analysis of the three-dimensional TPM configurations based on the screw theory. Lu et al. [Erreur ! Source du renvoi introuvable.] proposed a 3-RRRP (4R) three-translation PM and analyzed the kinematics and workspace. Yang et al. [Erreur ! Source du renvoi introuvable.] studied 3T0R PMs based on the single opened chains (SOC) units, wherein a variety of new TPMs were synthesized and then classified [16]. Chablat et al. [17] proposed a Cartesian TPM, i.e., Orthoglide, driven by prismatic joints. After that, Pashkevish et al. [18] performed the kinematics and workspace analysis of this mechanism. Zeng et al. [19-21] designed a prismatic joints-actuated three-translational Tri-pyramid PM, and designed a varieties linearly actuated variants, with their kinematic analysis. Prause et al. [22] compared the characteristics of a family of linearly driven 3-DOF TPMs with respect to dimensional synthesis, boundary conditions, and workspace, for the selection of mechanisms with better performances. Jha et al. [23] analyzed the singularity and workspace of four Delta-like 3-DOF TPMs. Shen et al.[24] studied a 3−DOF translational PM with partial motion decoupling and analytic direct kinematics, and analyzed its conditions of the singular configurations.

---

[1] Throughout this paper, P, R, U and C stand for prismatic, revolute, universal and cylindrical joints, respectively.

Most of the previously reported TPMs have the fully symmetrical topological architectures, leads to the highly nonlinear kinematic models due to the coupled input-output motion, introducing the difficulties in the motion control and trajectory planning. On the other hand, the asymmetric architecture can ensure advantages of motion decoupling for mechanisms. Moreover, the parallelogram (Pa, a.k.a, Π joint) structure is an important linkage to lay out TPMs, while, this introduces the structural complexity due to the presence of the closed sub-loop. Thus, the fewer use of parallelogram can ease the structural complexity in turn. Taking into consideration the two previous aspects, the design of TPMs with decoupled motion and lower complexity will be the focus of this paper. Using a topological design method of parallel mechanism (PM) based on position and orientation characteristic (POC) equations, this paper presents a novel three-degree-of-freedom (3-DOF) TPM. The TPM can be used for manufacturing large work pieces when the actuated joints move along a long-distance guide rail.

The remaining of the paper is organized as follows. In Section 1, the kinematic structure of a novel TPM is proposed referring to the topology design theory of PM based on position and orientation characteristics (POC) equations [25, 26]. This 3-DOF TPM includes fewer joints and links compared with Delta robot. Section 2 deals with an effective kinematic modeling of the TPM based on its topological characteristics, in context of POC, degree of freedom and coupling degree, from which its symbolic direct kinematic solutions are readily obtained, together with the evaluation of decoupled properties of the input-output motion without Jacobin analysis. In Section 3, with aid of the inverse kinematic solutions, the singular configurations of the parallel manipulator are identified, wherein the corresponding singular surfaces are graphically presented by using a Gröbner based elimination operation in Maple software. Section 4 presents the evaluation on the workspace of the proposed parallel manipulator by means of the 3D visualization. Finally, conclusions are drawn in the last part.    .

## 1. Topological design

Figure 1 depicts the proposed 3-DOF TPM, of which the base platform 0 is connected to the moving platform 1 by two hybrid chains that contain closed loop(s) and joints in serial. The structural and geometric constraints of the two hybrid chains (HC) of the TPM are given in such a way below.

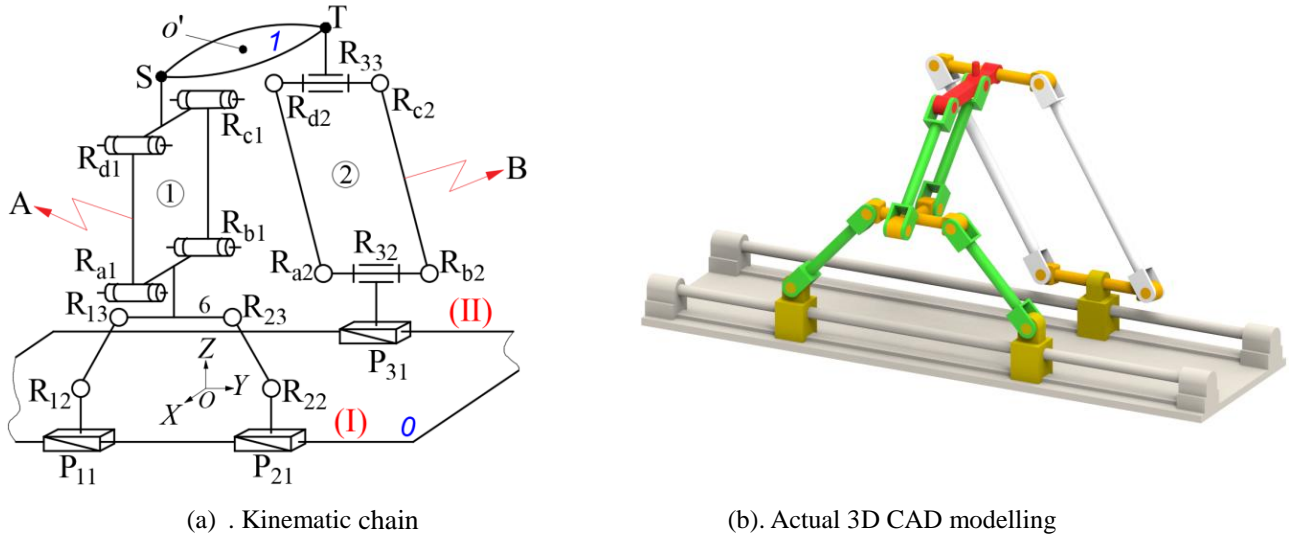

(a) . Kinematic chain             (b). Actual 3D CAD modelling

Figure 1. The proposed 3-DOF TPM

1) For the hybrid chain guided by linear rail (I), the six-bar planar mechanism loop (i.e., $P_{11}R_{12}R_{13}R_{23}R_{22}P_{21}$, denoted as 2P4R planar mechanism) with the motions confined in a plane that is perpendicular to another plane in which the motions of 4R parallelogram mechanism① (i.e., $R_{a1}R_{b1}R_{c1}R_{d1}$, shortly written as $\Diamond(a_1b_1c_1d_1)$, and denoted as Pa①) exist. These two linkages are arranged in serial to connect the moving platform 1 at point $O'$ to form the hybrid chain $A$ ($HC_A$). It is noted that the axes of the revolute joints $R_{12}$, $R_{13}$, $R_{23}$, $R_{22}$ and $R_{a1}$, $R_{b1}$, $R_{c1}$, $R_{d1}$ are all parallel to the X-axis. Two P-joints $P_{11}$ and $P_{21}$ of the six-bar planar mechanism are seen as the actuated joints.

2) For the hybrid chain moving along linear guide rail (II), the actuated P-joint $P_{31}$ rigidly connects to R-joint $R_{32}$ and a 4R parallelogram mechanism②(i.e., $R_{a2}R_{b2}R_{c2}R_{d2}$, shortly written as $\Diamond(a_2b_2c_2d_2)$, and denoted as Pa②) in serial, of which the link end in the Pa② is connected to the moving platform 1 at point T by revolution joint $R_{33}$ to form the hybrid chain $B$ ($HC_B$).

3) The linear guide rails (I) and (II) are parallel. When the TPM moves, link 6 is the output element of the six-bar planar mechanism with two translations that will be proved in a plane normal to the base platform plane.

This novel TPM can have three advantages below.

①The numbers of total joints and components of the proposed mechanism equal to 17 and 16, respectively, while for Delta robot, the corresponding numbers are equal to 21 and 17. Therefore, compared with the well-known Delta robot, the proposed TPM has is simpler structure for rapid prototyping.

②The guide rails to support the actuated P-joints can have finite lengths according to actual needs, which ensures larger workspace than Delta robot, to extend its industrial applications.

③The TPM has symbolic solutions to the direct and inverse kinematic problems and partially decoupled motion that is verified in the following sections. The symbolic direct position solution is beneficial for error analysis, workspace analysis, and velocity/acceleration and dynamics analysis. Moreover, partial motion decoupling is helpful to trajectory planning and mechanism control. Regarding these two issues, the proposed TPM outperforms in operation compared to Delta robot.

## 2. Kinematics modelling method based on topological characteristics

In this section, an effective kinematic modeling method of TPM based on its topological characteristics, with respect to the POC, DOF and coupling degree, is proposed, from which the symbolic direct kinematic solutions are easily obtained, and the decoupled properties of input-output motion are also directly evaluated without Jacobian calculation and analysis.

### 2.1 The kinematics modeling method based on topological characteristics

The basic problem-solving idea of the kinematics modeling method based on topological characteristics is depicted as follows. Firstly, according to the mechanism topological decomposition method, a PM can be decomposed into several sub-kinematic chains(SKC) with a coupling degree of $k_i$. Meantime, the PM can be grouped into two categories, namely, the PM with only one SKC and the PM with multiple SKCs. Secondly, according to the availability of topological characteristics of the PM, the position equations of each $SKC_i$ can be established one after another. Finally, the algebraic method or numerical method is applied to solve the position equations, and the symbolic solutions or closed-form solutions or numerical solutions of these problems of three types can be obtained, respectively. In detail, this method can be implemented in two steps:

1) *Topological analysis*[27]. In this step, the first two important topological characteristics, i.e., position and orientation characteristic (POC) and degree of freedom(DOF) of the PM can be obtained. In addition, the PM is decomposed into a series of single-open-chain (SOC) with constraint degree of three types, i.e., positive, zero and negative one, then the constraint degree value ($\Delta$) of each SOC can be obtained. These SOCs can be further divided into several sub-kinematic-chains (SKC), and the coupling degree $k_i$ of each SKC is calculated ($k_i = \Delta_{+j} = |\Delta_{-j}|$). Now the PM that contains only one SKC or multiple SKCs can be determined.

2) *Establishment and solutions of the position equation*. According to the availability of the topological characteristics of the PM, the position equations of each $SKC_i$ can be easily established. Since the position equation of each $SKC_i$ is simple, conventional mathematical methods can be applied to calculate the direct solutions of these SKCs and then the whole PM.

In the following, the kinematics modelling method based on topological characteristics are illustrated in two steps. The first step is to analyze the topology characteristics of the PM. The next one lies in position analysis in terms of both direct and inverse kinematic solutions. In this kinematic modeling method, the number of virtual variable is assigned according to the coupling degree obtained from topological analysis.

**2.2 Analysis of topological characteristics**

*2.2.1 Analysis of the POC set*

The POC equations for serial and parallel mechanisms are expressed respectively as follows[26]:

$$M_{bi} = \bigcup_{i=1}^{m} M_{Ji} \tag{1}$$

$$M_{Pa} = \bigcap_{i=1}^{n} M_{bi} \tag{2}$$

where

$M_{Ji}$ - POC set generated by an $i$-th joint.

$M_{bi}$ - POC set generated by the end link of $i$-th branched chain.

$M_{Pa}$ - POC set generated by the moving platform of PM.

POC- position and orientation characteristics

∪ -union operation

∩ -intersection operation

Accordingly, the output motions of the intermediate link 6 in the 2P4R six-bar planar mechanism inside the hybrid chain $A$ include two translations and one rotation (2T1R), which is denoted by

$$\begin{bmatrix} t^2(\perp R_{13}) \\ r^1(\parallel R_{13}) \end{bmatrix}$$

Where $t^2(\perp R_{13})$ means that there are two translations lying in a plane normal to the axis of joint $R_{13}$, and, $r^1(\parallel R_{13})$ means that there is one rotation with the axis of rotation parallel to the axis of joint $R_{13}$. The other notations in the formulas above can be found in [26,27].

Therefore the output motions of the end link of the hybrid chain $A$ (denoted as $HC_A$) turns out to three-translation and one-rotation (3T1R), since the output motion of the Pa①is 1-DOF translation. Similarly, the output motions of hybrid chain $B$ (denoted as $HC_B$) are three-translation and one-rotation (3T1R) too.

Thus, the POC sets of the end of the two hybrid chains $HC_A$ and $HC_B$ are determined according to *Eqs.* (1) and (2) as follows.

$$M_{HC_A} = \begin{bmatrix} t^2(\perp R_{13}) \\ r^1(\parallel R_{13}) \end{bmatrix} \cup \begin{bmatrix} t^1(\parallel \Diamond(a_1b_1c_1d_1)) \\ r^0 \end{bmatrix} = \begin{bmatrix} t^3 \\ r^1(\parallel R_{13}) \end{bmatrix}$$

$$M_{HC_B} = \begin{bmatrix} t^3 \\ r^1(\parallel R_{32}) \end{bmatrix} \cup \begin{bmatrix} t^1(\parallel \Diamond(a_2b_2c_2d_2)) \\ r^0 \end{bmatrix} = \begin{bmatrix} t^3 \\ r^1(\parallel R_{32}) \end{bmatrix}$$

As a consequence, the POC set of the moving platform 1 of this TPM is determined from *Eq.* (2) below

$$M_{Pa} = M_{HC_A} \text{ I } M_{HC_B} = \begin{bmatrix} t^3 \\ r^0 \end{bmatrix}$$

This formula implies that the hybrid chain $HC_B$ constrains the rotational output of link 6 of six-bar planar mechanism around the axis of joint $R_{13}$ in the hybrid chain $HC_A$, which leads to *Eq*. (8) in Section 2.3.2. This is very important topological constraint property of the TPM, which will significantly simplifies the process of the direct solutions that are described in the last paragraph of Section 2.2.3.

Moreover, the hybrid chain *A* also constrains the rotational output of the end of the hybrid chain *B* around the axis of joint $R_{33}$ too. Therefore, the moving platform 1 of the TPM can generate only three translational motions.

*2.2.2 Determining the DOF*

The general and full-cycle DOF formula for PMs proposed in [26] is given by

$$F = \sum_{i=1}^{m} f_i - \sum_{j=1}^{v} \xi_{Lj} \tag{3}$$

$$\sum_{j=1}^{v} \xi_{Lj} = \dim \left\{ (\underset{i=1}{\overset{j}{\text{I}}} M_{b_i}) Y M_{b_{(j+1)}} \right\} \tag{4}$$

where

*F* - DOF of PM.

$f_i$ - DOF of the $i^{th}$ joint.

*m* - number of all joints of the PM.

*v* - number of independent loops of the PM, and *v=m-n+1*.

*n* - number of links.

$\xi_{L_j}$ - number of independent displacement equations of the $j^{th}$ loop.

$\underset{i=1}{\overset{j}{\text{I}}} M_{b_i}$ - POC set generated by the sub-PM formed by the former *j* branches.

$M_{b(j+1)}$ - POC set generated by the end link of *j+1* sub-chains.

The TPM can be decomposed into two independent loops, and their independent displacement equations are expressed as follows:

①The first loop, denoted as $LOOP_1$, consists of the 2P4R planar mechanism in the hybrid chain *A*. Thus, the number of independent displacement equation of the 2P4R planar mechanism is $\xi_{L_1} = 3$.

②The aforementioned 2P4R planar mechanism and the 4R parallel mechanism Pa①, plus hybrid chain *B* will form the second loop, denoted as $LOOP_2$. According to *Eq*. (4), the number of independent displacement equation $\xi_{L_2}$ of the second loop can be obtained as below.

$$\xi_{L2} = \dim \left\{ \begin{bmatrix} t^2(\perp R_{13}) \\ r^1(\parallel R_{13}) \end{bmatrix} Y \begin{bmatrix} t^1(\parallel \Diamond(a_1b_1c_1d_1)) \\ r^0 \end{bmatrix} Y \begin{bmatrix} t^3 \\ r^1(\parallel R_{32}) \end{bmatrix} \right\} = \dim \left\{ \begin{bmatrix} t^3 \\ r^2(\parallel \Diamond(R_{13}, R_{32})) \end{bmatrix} \right\} = 5$$

Thus, the DOF of the TPM is calculated from *Eq*. (3) expressed as

$$F = \sum_{i=1}^{m} f_i - \sum_{j=1}^{v} \xi_{L_j} = (6+5)-(3+5) = 3$$

Therefore, the DOF of the TPM is equal to 3. The prismatic joints $P_1$, $P_2$ and $P_3$ on the base platform 0 are selected as the actuated joints.

### 2.2.3 Determining the coupling degree

According to the mechanism composition principle based on single-opened-chains (SOC) units[25,26], any PM can be decomposed into several sub-kinematics chains (SKC), and a SKC with $v$ independent loops can be further decomposed into $v$ SOC. The constraint of the $j^{th}$ SOC, $\Delta_j$, is defined by[26]

$$\Delta_j = \sum_{i=1}^{m_j} f_i - I_j - \xi_{L_j} = \begin{cases} \Delta_j^- = -5,-4,-2,-1 \\ \Delta_j^0 = 0 \\ \Delta_j^+ = +1,+2,+3,\cdots \end{cases} \quad (5)$$

Where

$\Delta_j$ - constraint degree of the $j^{th}$ SOC.

$m_j$ - number of joints contained in the $j^{th}$ SOC$_j$.

$I_j$ - number of actuated joints in the $j^{th}$ SOC$_j$.

$f_i$, $\xi_{L_j}$ - the same definition as in *Eqs*.(3) and (4).

For a SKC, it must be satisfied with the following equation.

$$\sum_{j=1}^{v} \Delta_j = 0 \quad (6)$$

Sequentially, the coupling degree of a SKC is then defined as[25,26]

$$\kappa = \frac{1}{2} min\left\{\sum_{j=1}^{v} |\Delta_j|\right\} \quad (7)$$

The physical meaning of the coupling degree $\kappa$ can be interpreted in this way. The coupling degree $k$ describes the complexity of the topological structure of a PM, which can indicate that the kinematic and dynamic analysis is complex or not. It has been proved that the higher the coupling degree $\kappa$ is, the more complex the kinematic and dynamic solutions of the PM are [26, 27].

The number of the independent displacement equations of $LOOP_1$ and $LOOP_2$ have been calculated in Section 2.2.2, i.e., $\xi_{L_1} = 3$, $\xi_{L_2} = 5$, thus, the constraint degree of the two independent loops are calculated by *Eq*. (5), respectively, with the solution below:

$$\Delta_1 = \sum_{i=1}^{m_1} f_i - I_1 - \xi_{L_1} = 6-2-3 = 1$$

$$\Delta_2 = \sum_{i=1}^{m_2} f_i - I_2 - \xi_{L_2} = 5-1-5 = -1$$

Thus, according to *Eq*. (6), the TPM contains only one SKC. The coupling degrees of the SKC is calculated by *Eq*. (7) as

$$k = \frac{1}{2}\sum_{j=1}^{v}|\Delta_j| = \frac{1}{2}(|+1|+|-1|) = 1$$

In general, when solving the direct position solutions of the PM, only one virtual variable will be assigned in the first loop whose constraint degree is positive one $(\Delta_j=+1)$. Then, one constraint equation with this virtual variable is established in the second loop with the negative constraint degree $(\Delta_j=-1)$. Further, the real value of this virtual variable can be obtained by the one-dimensional numerical method, thus, the direct position solutions of the PM are obtained finally.

However, the proposed TPM is a special case of 3-translational PM, the second loop with the negative constraint degree $(\Delta_j=-1)$ directly exerts a special topological constraint to the first loop with the positive constraint degree is one $(\Delta_j=+1)$, which means that the output motion of the link 6 in the first loop is always parallel to plane that the base platform 0 lies in. Therefore, the virtual variable is easily obtained from the first loop directly, and there is no need to solve the virtual variable by one-dimensional numerical method, which significantly simplifies the process of the direct solutions. Thus, the symbolic direct position solutions of the TPM can be directly obtained as depicted in the following Section 2.3.2.

**2.3 Position analysis**

**2.3.1 The coordinate system and parameterization**

The kinematics architecture of the TPM is shown in Fig.2. The base platform 0 is rectangular with a width of $2b$. The global coordinate system O-XYZ is established on the base platform 0, with the origin located at the geometric center. The X-axis and Y-axis are perpendicular, parallel to segment $A_1A_2$, respectively, and the Z-axis is normal to the base plane pointing upwards. The moving coordinate system O'-X'Y'Z' is established with the coordinate axes parallel to those of the global frame with the origin located at the geometric center O' of the moving platform 1.

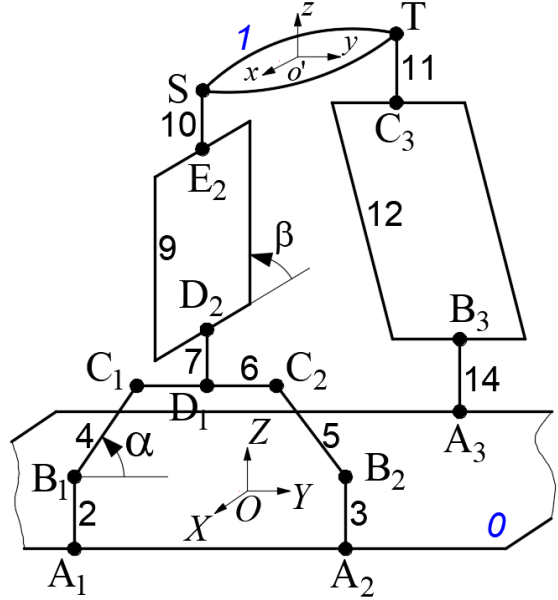

Figure 2. Kinematic modelling of the TPM

Let $A_1B_1=A_2B_2=A_3B_3=l_1$, $B_1C_1=B_2C_2=l_2$, $B_3C_3=l_9$, $C_1C_2=l_3$, $D_1D_2=l_4$, $D_2E_2=l_6$, $E_2S=l_7$, $ST=2d$, $TC_3=l_8$

In order to simplify the structure and improve the kinematic and dynamic performance of the TPM, we let $D_1D_2=l_4=0$, $E_2S=l_7=0$ and $TC_3=l_8=0$ when prototyping, such as the actual 3D model shown in Fig. 1(b).

Further, the angle between the vectors $B_1C_1$ and the Y-axis is $\alpha$, and let $\alpha$ be the virtual variable. The angle between the vectors $D_2E_2$ and the negative X-axis is $\beta$.

### 2.3.2 Direct kinematic problem

For the direct kinematics, the values of the actuated joint variables ($y_{A1}$, $y_{A2}$, $y_{A3}$) are known, and the objective is to find the tool center point location $O'(x,y,z)$ that is determined by the given joint parameters.

**1) Solving the first loop**

$$LOOP_1 : A_1 - B_1 - C_1 - C_2 - B_2 - A_2$$

The coordinates of points $A_i$ and $B_i$ ($i=1, 2, 3$) on the base platform 0 are derived, respectively

$$A_1 = (b, y_{A_1}, 0)^T, A_2 = (b, y_{A_2}, 0)^T, A_3 = (-b, y_{A_3}, 0)^T; \quad B_1 = (b, y_{A_1}, l_1)^T, B_2 = (b, y_{A_2}, l_1)^T, B_3 = (-b, y_{A_3}, l_1)^T$$

Since the output link 6 of the 2P4R planar mechanism is always parallel to the base platform 0 without rotational output, as described in the last paragraph of Section 2.2.1, we have $C_1C_2 \| A_1A_2$. Then the following constraint equation is derived.

$$z_{C_1} = z_{C_2} \tag{8}$$

Therefore, the coordinates of points $C_1$ and $C_2$ are calculated as

$$C_1 = (b, y_{A_1} + l_2 \cos\alpha, l_1 + l_2 \sin\alpha)^T, \quad C_2 = (b, y_{A_1} + l_2 \cos\alpha + l_3, l_1 + l_2 \sin\alpha)^T$$

With the link length constraints defined by $B_2C_2=l_2$, two constraint equations can be deduced as below,

$$B^2 - 2Bl_2 \cos\alpha = 0$$

Thus

$$\cos\alpha = B/2l_2, \quad \text{while } \sin\alpha = m\sqrt{1-\cos^2\alpha} \tag{9}$$

with

$$B = y_{A2} - y_{A1} - l_3$$

It is noteworthy that the constraint equation $z_{C_1} = z_{C_2}$ is the key to directly solving the virtual variable α, leading to the symbolic direct position solutions.

**2) Solving the second loop**

$$LOOP_2 : D_1 - D_2 - E_2 - S - T - C_3 - B_3 - A_3$$

The coordinates of the points $D_1$ and $D_2$ obtained from points $C_1$ and $C_2$ are calculated as

$$D_1 = (b, y_{A_1} + l_2 \cos\alpha + l_3/2, l_1 + l_2 \sin\alpha)^T, \quad D_2 = (b, y_{A_1} + l_2 \cos\alpha + l_3/2, l_1 + l_2 \sin\alpha + l_4)^T$$

The coordinates of the points $E_2$ can be obtained directly

$$E_2 = \begin{bmatrix} b - l_6 \cos\beta \\ y_{A_1} + l_2 \cos\alpha + l_3/2 \\ l_1 + l_2 \sin\alpha + l_4 + l_6 \sin\beta \end{bmatrix}$$

For the calculation of the coordinates of point $O'$:

$$o' = \begin{bmatrix} x \\ y \\ z \end{bmatrix} = \begin{bmatrix} b - l_6 \cos\beta - d \\ y_{A_1} + l_2 \cos\alpha + l_3/2 \\ l_1 + l_4 + l_7 + l_2 \sin\alpha + l_6 \sin\beta \end{bmatrix} \qquad (10)$$

Further, the coordinates of point $C_3$ are represented with the known coordinates of point $O'$ as below:

$$C_3 = (x - d, y, z - l_8)^{\mathrm{T}}$$

With the link length constraints defined by $C_3B_3 = l_9$, the constraint equation can be deduced:

$$G_1 \sin\beta + G_2 \cos\beta + G_3 = 0$$

resulting in the solutions below

$$\beta = 2\arctan\frac{-G_1 + n\sqrt{G_1^2 + G_2^2 - G_3^2}}{G_3 - G_2} \qquad (11)$$

where

$$G_1 = 2F_3 l_6, \quad G_2 = -2F_1 l_6, \quad G_3 = F_1^2 + F_2^2 + F_3^2 + l_6^2 - l_9^2;$$

$$F_1 = 2b - 2d, \quad F_2 = y_{A_1} - y_{A_3} + l_2 \cos\alpha + l_3/2, \quad F_3 = l_4 + l_7 - l_8 + l_2 \sin\alpha$$

Consequently, by substituting the values of α and β obtained from *Eqs.* (9) and (11) into *Eq.* (10), the coordinates of point $O'$ in the global coordinate system can be obtained, namely,

$$\begin{cases} x = f_1(y_{A_1}, y_{A_2}, y_{A_3}) \\ y = f_2(y_{A_1}, y_{A_2}) \\ z = f_2(y_{A_1}, y_{A_2}, y_{A_3}) \end{cases} \qquad (12)$$

Therefore, from *Eq.*(12), the TPM not only has symbolic solutions to the direct kinematic problems, but also has input-output partial motion decoupling property, i.e., the $y$ value at the output is determined only by the inputs $y_{A1}$ and $y_{A2}$. This is very useful for trajectory planning and motion control of moving platforms. Here, the coefficients *m* and *n*, respectively, correspond to two different values, *i.e.*, $m = \pm 1$ and $n = \pm 1$. Therefore the number of the direct position solutions is equal to 4.

The preceding calculation process can be expressed in Fig.3.

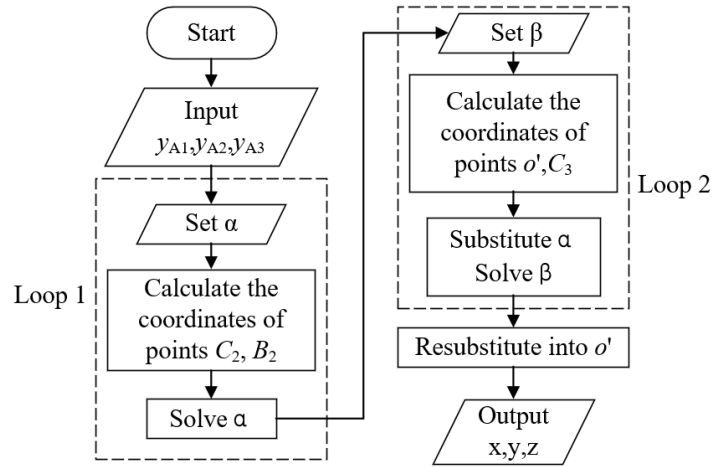

Figure 3. Calculation of direct position solutions.

### 2.3.3 Inverse kinematic problem

For the inverse kinematics, the position of the end-point O'(x, y, z) is known, and the objective is to find the actuated joint variables ($y_{A1}$, $y_{A2}$, $y_{A3}$) that yield the given location of the tool.

The coordinates of the points $E_2$ and $C_3$ can be obtained from the point O'.

$$E_2 = (x+d, y, z-l_7)^T ; \qquad C_3 = (x-d, y, z-l_8)^T ;$$

Thus, the coordinates of points $C_1$ and $C_2$ are defined from that of point $D_2$ and $D_1$ in sequence as:

$$D_2 = (x+d+l_6\cos\beta, y, z-l_7-l_6\sin\beta)^T, \qquad D_1 = (x+d+l_6\cos\beta, y, z-l_7-l_6\sin\beta-l_4)^T$$

$$C_1 = (x+d+l_6\cos\beta, y-l_3/2, z-l_7-l_6\sin\beta-l_4)^T$$

$$C_2 = (x+d+l_6\cos\beta, y+l_3/2, z-l_7-l_6\sin\beta-l_4)^T$$

The x coordinates of points $D_2$, $D_1$, $C_2$, and $C_1$ are all equal to b, thus the cosine of the $\beta$ angle can be obtained as.

$$\cos\beta = \frac{b-d-x}{l_6}; \quad \sin\beta = v\sqrt{1-\cos^2\beta} \tag{13}$$

Therefore, with the length constraints defined by $B_1C_1=l_2$, $B_2C_2=l_2$ and $B_3C_3=l_9$, there are three constraint equations below.

$$\begin{aligned}
(x_{C1}-x_{B1})^2 + (y_{C1}-y_{B1})^2 + (z_{C1}-z_{B1})^2 &= l_2^2 \\
(x_{C2}-x_{B2})^2 + (y_{C2}-y_{B2})^2 + (z_{C2}-z_{B2})^2 &= l_2^2 \\
(x_{C3}-x_{B3})^2 + (y_{C3}-y_{B3})^2 + (z_{C3}-z_{B3})^2 &= l_9^2
\end{aligned} \tag{14}$$

From *Eq*s. (14), $y_{Ai}$ (i=1, 2, 3) are calculated as following

$$y_{A_i} = M_i + w\sqrt{-L_i - N_i}, i = 1,2,3 \tag{15}$$

where

$$L_1 = L_2 = (x+d+l_6\cos\beta-b)^2, L_3 = (x-b+d)^2$$

$$M_1 = y - l_3/2, M_2 = y + l_3/2, M_3 = y$$

$$N_1 = N_2 = (z-l_7-l_6\sin\beta-l_4-l_1)^2 - l_2^2, N_3 = (z-l_8-l_1)^2 - l_9^2$$

In summary, when the coordinates of point O' on the moving platform 1 are known, v and w correspond to two different values, i.e., $v=\pm 1$ and $w=\pm 1$. Thus, each input value $y_{Ai}$ (i=1, 2, 3) has two sets of solutions. Therefore, the number of the inverse position solutions is equal to 16.

### 2.3.4 Numerical simulation for direct and inverse kinematics

To verify the correctness of the direct and inverse kinematics models, the structural parameters of the TPM are given in unit of millimeter as shown in Table 1.

Table 1. The structural parameters of the TPM.

| $a$ | $b$ | $d$ | $l_1$ | $l_2$ | $l_3$ | $l_4$ | $l_5$ | $l_6$ | $l_7$ | $l_8$ | $l_9$ | $l_{10}$ |
|---|---|---|---|---|---|---|---|---|---|---|---|---|
| 360 | 90 | 45 | 70 | 160 | 120 | 0 | 90 | 180 | 0 | 0 | 300 | 150 |

**1) Direct solutions**

According to the structural parameters given in Table 1, the CAD model of the TPM is shown in Fig.1(b). A set of input values and the corresponding output values measured from the actual 3D model are $y_{A1}$ = -111.24, $y_{A2}$ = 244.70, $y_{A3}$ = 246.92, and $x$ = -80.39, $y$ = 66.73, $z$ = 307.23 (mm), respectively.

These parameters are substituted into *Eqs*. (9) to (11), and the corresponding four direct solutions are calculated, as shown in Table 2, for which the four corresponding configurations are shown in Fig. 4(a)~(d).

Table 2. Theoretical calculation of direct kinematic solution.

| NO. | $m$ | $n$ | $x$ | $y$ | $z$ |
|---|---|---|---|---|---|
| 1* | +1 | +1 | -80.3862 | 66.7300 | 307.2328 |
| 2 | +1 | -1 | 194.7183 | 66.7300 | 78.1662 |
| 3 | -1 | +1 | 194.7183 | 66.7300 | 61.8338 |
| 4 | -1 | -1 | -80.3862 | 66.7300 | -167.2328 |

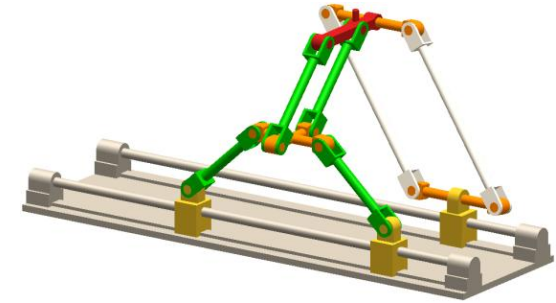

(a) *m*=+1, *n*=+1

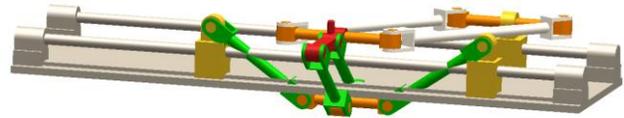

(b) *m*=+1, *n*=-1

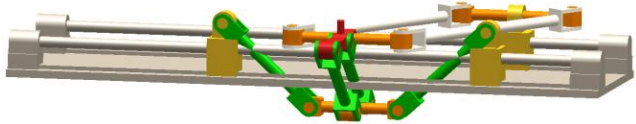

(c) *m*=-1, *n*=+1

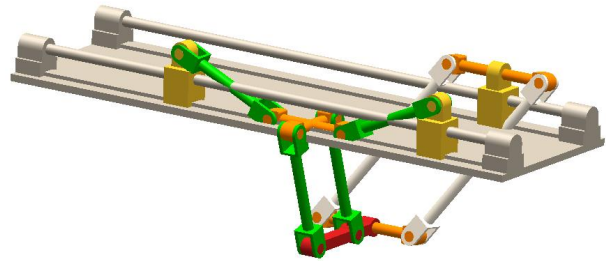

(d) *m*=-1, *n*=-1

Figure 4. The four configurations corresponding to the four direct solutions

Removing the mechanism architectures of mechanical collision, only the first configuration exists, which means the direct kinematic solution are correct. So the coefficient *m* equals +1 (see *Eq*.(9)), and *n* equals +1(see *Eq*.(11)).

## 2) Inverse solution

Substituting the measured end position values of the direct position solution into *Eqs*. (13) and (15), when the sine value of the β angle is negative ($v=-1$), the corresponding input value is a complex number, so there are only 8 sets of inverse solutions exist, as shown in Table 3. It can be seen from the structure of the TPM that in the positive direction of the Y axis, the value of active joint $y_{A2}$ is always greater than the value of $y_{A1}$. Therefore, the third and fourth cases do not exist mechanically. Thus, there are only six actual inverse solutions.

Table 3. Theoretical calculation of inverse kinematic solutions.

| No. | 1 | 2 | 3 | 4 | 5* | 6 | 7 | 8 |
|---|---|---|---|---|---|---|---|---|
| $y_{A1}$ | 124.6992 | 124.6992 | 124.6992 | 124.6992 | -111.2392 | -111.2392 | -111.2392 | -111.2392 |
| $y_{A2}$ | 244.6992 | 244.6992 | 8.7608 | 8.7608 | 244.6992 | 244.6992 | 8.7608 | 8.7608 |
| $y_{A3}$ | 246.9229 | -113.4629 | 246.9229 | -113.4629 | 246.9229 | -113.4629 | 246.9229 | -113.4629 |

Fig.5(a)-5(f) shows the six actual configurations corresponding to the six correct inverse solutions in Table 3.

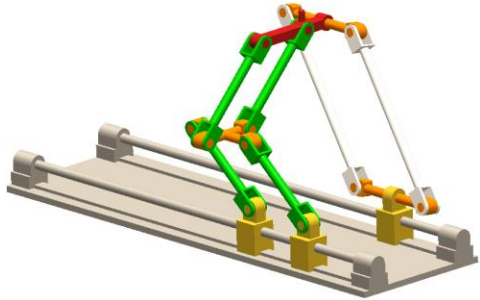
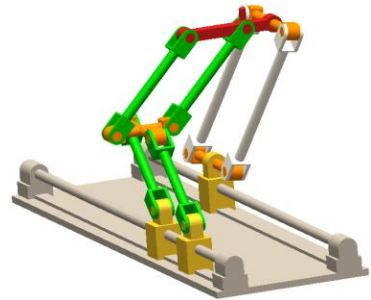

(a) No.1　　　　　　　　　　　　　　　　(b) No.2

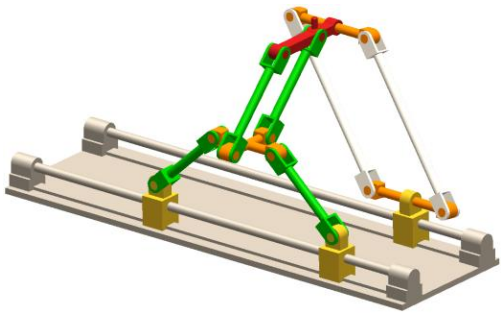
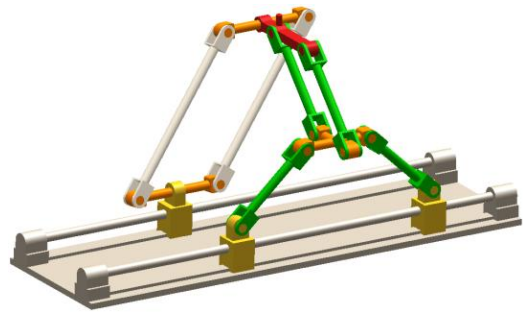

(c) No.5　　　　　　　　　　　　　　　　(d) No.6

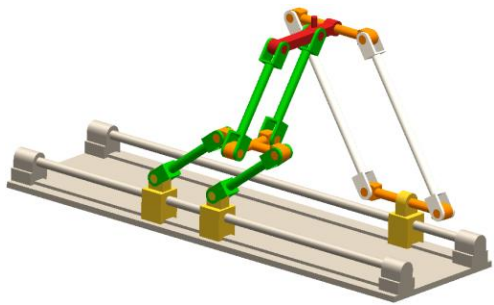
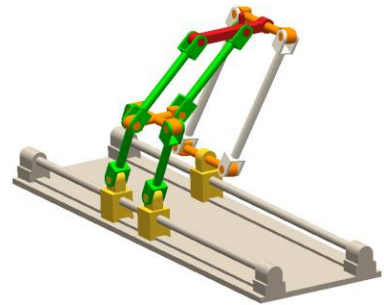

(e) No.7　　　　　　　　　　　　　　　　(f) No.8

Figure 5. The six actual configurations corresponding to the six correct inverse solutions

Among these six inverse solutions, the observations are given as follows.

(1) The fifth inverse solution in Table 3 is consistent with the measured values of the three inputs. Similarly, the other forward solution data in Table 1 also verifies the correctness of the forward and inverse solution formulas.

(2) Only the fifth and sixth sets corresponding Fig.5 (c), (d) respectively are non singular solutions. The fifth set of values corresponds to the value of the TPM taken in 3D model shown in Fig.1 (b).

(3) The first, second, seventh and eighth solutions are singular configurations, shown in Fig.5 (a), (b), (e) and (f) respectively, which will be explained by Fig.7 (a) corresponding the Case ①of parallel singular configuration in the following section.

## 3. Singularity analysis

When the TPM is in a singular configuration, its motion is undetermined. The singular configurations can be identified by Jacobian analysis of the robotic mechanism.

### 3.1 Jacobian matrix

By differentiating *Eq.* (14) with respect to time, the mapping between the output velocity of the end-effector of the moving platform $\dot{x} = [\dot{x} \ \dot{y} \ \dot{z}]^T$ and the input velocity of the actuated joints $\dot{q} = [\dot{q}_{A1} \ \dot{q}_{A2} \ \dot{q}_{A3}]^T$ is obtained as

$$A\dot{x} + B\dot{q} = 0 \qquad (16)$$

where

$$A = \begin{bmatrix} f_{11} & f_{12} & f_{13} \\ f_{21} & f_{22} & f_{23} \\ f_{31} & f_{32} & f_{33} \end{bmatrix}, \quad B = diag(u_{11}, u_{22}, u_{33})$$

$f_{11} = \cot\beta(z_{C1} - z_{B1})$; $f_{12} = -(y_{C1} - y_{B1})$; $f_{13} = -(z_{C1} - z_{B1})$;

$f_{21} = \cot\beta(z_{C2} - z_{B2})$; $f_{22} = -(y_{C2} - y_{B2})$; $f_{23} = -(z_{C2} - z_{B2})$;

$f_{31} = x_{C3} - x_{B3}$; $f_{32} = y_{C3} - y_{B3}$; $f_{33} = z_{C3} - z_{B3}$;

$u_{11} = (y_{C1} - y_{B1})$; $u_{22} = (y_{C2} - y_{B2})$; $u_{33} = -(y_{C3} - y_{B3})$;

### 3.2 Singular configurations analysis

In *Eq.* (16), **A** and **B** are named as the parallel and serial Jacobian matrices, respectively. The serial singularities occur if det(**B**) = 0 and the parallel singularities occur whenever det(**A**) = 0.

#### 3.2.1 Serial singularity

From det (B) = 0, the set of determinant solutions of matrix B is:

$$u = \{u_{11} \ Y \ u_{22} \ Y \ u_{33}\}$$

where

$u_{11} = \{y_{C1} - y_{B1} = 0\}$, which means link $B_1C_1$ is parallel to the Z-axis.

$u_{22} = \{y_{C2} - y_{B2} = 0\}$, which means link $B_2C_2$ is parallel to the Z-axis.

In particular, when link $B_1C_1$ is parallel to the Z-axis, link $B_2C_2$ is also parallel to the Z-axis, that is, the above two conditions will occur simultaneously.

$u_{33} = \{y_{C3} - y_{B3} = 0\}$, which means the projections of link $A_3B_3$ and $B_3C_3$ on the YOZ plane are coincident, that is, the projections of $P_{31}R_{32}$ and $R_{32}R_{33}$ are coincident on the YOZ plane, as shown in Fig.6.

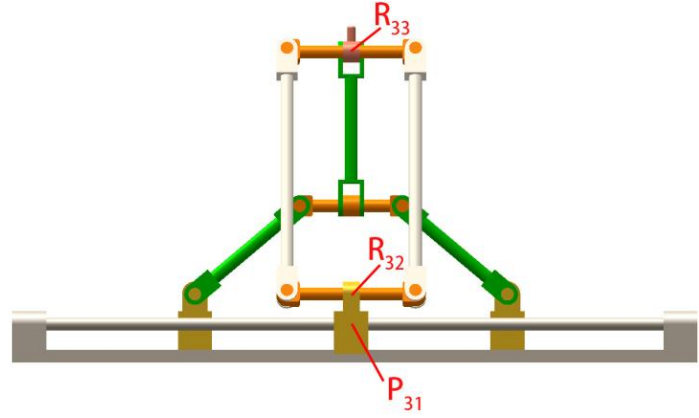

Figure 6. An example of serial singular configuration.

### 3.2.2 Parallel singularity

For the convenience of analysis, considering the matrix **A** as a combination of three-row vectors:

$$A = \begin{bmatrix} e_1 & e_2 & e_3 \end{bmatrix}^T$$

The determinant of matrix **A** Det(**A**) is equal to 0 when the two following situations occur:

(i) Two of the vectors are linearly related

① $e_1 = ke_2$ means that $e_1$ and $e_2$ are linearly related, thus, we obtain.

$$\frac{y_{C1} - y_{B1}}{z_{C1} - z_{B1}} = \frac{y_{C2} - y_{B2}}{z_{C2} - z_{B2}}$$

Therefore, the projections of $C_1B_1$ and $C_2B_2$ on the YOZ plane are parallel, that is, the projections of $R_{12}R_{13}$ and $R_{22}R_{23}$ on the YOZ plane are parallel, as shown in Fig.7 (a). This situation exactly corresponds to (2) and (7) in Figure 5.

② $e_1 = ke_3$ means that $e_1$ and $e_3$ are linearly related, yielding

$$\tan \beta = -\frac{z_{C3} - z_{B3}}{x_{C3} - x_{B3}}$$

Therefore, the slope of the projection of $C_3B_3$ on the XOZ plane is equal to the negative number of the tangent of the angle $\beta$, that is, the projections of $R_{b1}R_{c1}$ and $R_{32}R_{33}$ on the XOZ plane are parallel, which does not exist for the given lengths, as shown in Fig. 7 (b).

When $e_2 = ke_3$, the result is the same to the singularity of the case ②.

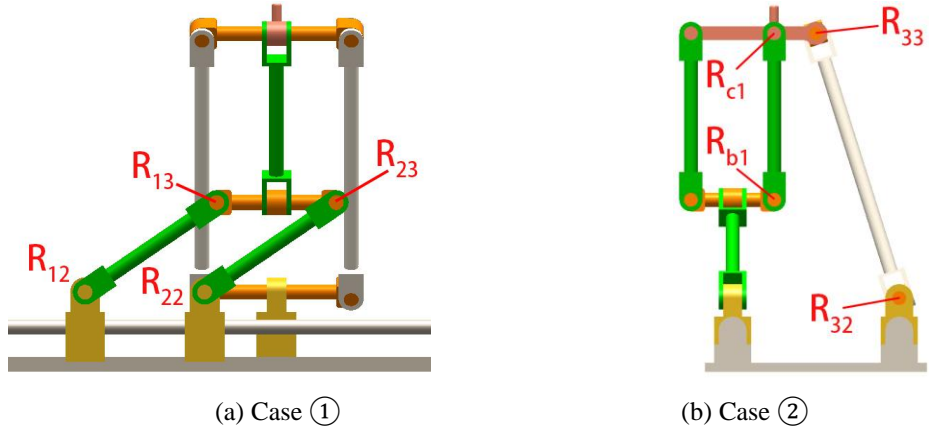

(a) Case ①          (b) Case ②

Figure 7. Examples of parallel singularity configuration.

(ii) Three vectors are linearly related

Let $e_2 = k_1 e_1 + k_2 e_3 (k_1 k_2 \neq 0)$, the following equation is obtained

$$\tan \beta = -\frac{z_{C3} - z_{B3}}{x_{C3} - x_{B3}}$$

Which means that the slope of the link $D_2 E_2$ is equal to the slope of the projection of $C_3 B_3$ on the XOZ plane, and the occurrence conditions are the same as those in Fig.7 (b).

### 3.3 Singularity surface analysis

Parallel and serial singularities as well as their projections in workspace and joint space can be computed using a Groebner based elimination method [4, 5]. This usual way for eliminating variables is to compute the algebraic closure of the projection of the parallel singularities in the workspace.

Based on the structure parameters given in Section 2.3.4, the singular surfaces inside the workspace and the joint space of the TPM are plotted with Maple as shown in Figs.8 and 9.

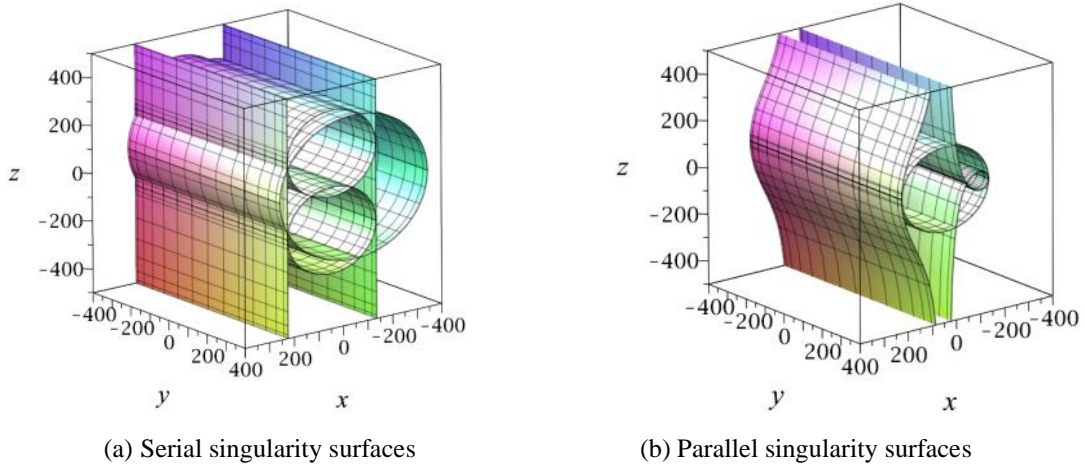

(a) Serial singularity surfaces        (b) Parallel singularity surfaces

Figure 8. Singularity surface in the mechanism workspace.

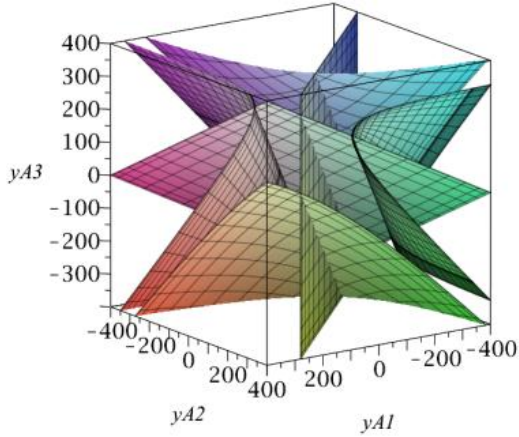 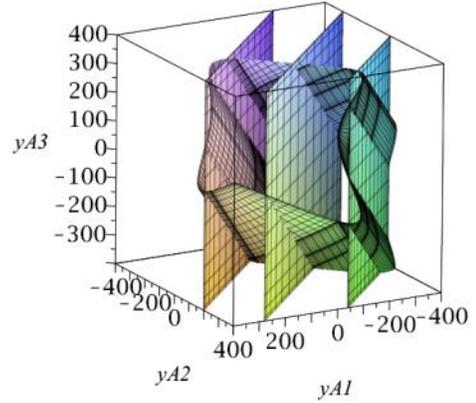

(a) Serial singularity surfaces  (b) Parallel singularity surfaces

Figure 9. Singularity surface in the joint space.

## 4. Workspace analysis

In this paper, the discrete method is adopted to determine the workspace of the TPM. First, by making use of the prescribed dimensions in Table 3, it is to find the searching space in the workspace. Then, based on the inverse kinematics solutions, and considering the constraints of the motion range of the actuated joints, the range of the passive joint, and the interference between the links, the workspace points that meet all the constraints are searched. The 3D image composed of these points is the workspace of the TPM. Finally, these spatial points are fitted to form a visual 3D workspace.

The search scope is set to:

$$-150 \leq x \leq 150, -200 \leq y \leq 200, 380 \leq z \leq 550$$

By using Matlab, the 3D workspace for the TPM is obtained, as shown in Fig.10 (a). Furthermore, the projection views of the workspace in the *yoz* direction and *xoz* direction are obtained, as shown in Fig. 10 (b) and Fig. 10(c).

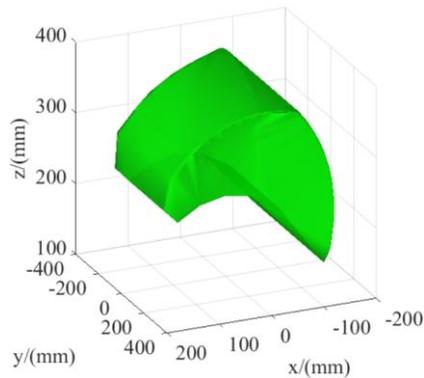

(a) 3D workspace

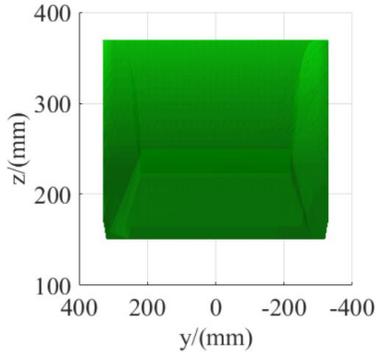
(b) Projection view of the *yoz* section

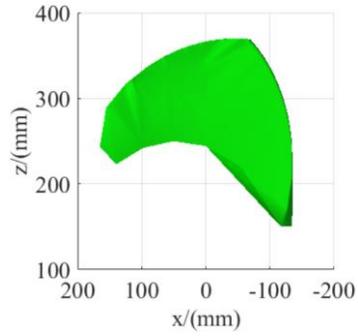
(c) Projection view of the *xoz* section

Figure 10. Workspace of the TPM.

By using Maple and SIROPA[4,5], the 3D workspace and the singularity surface are shown in one figure, as shown in Fig.11.

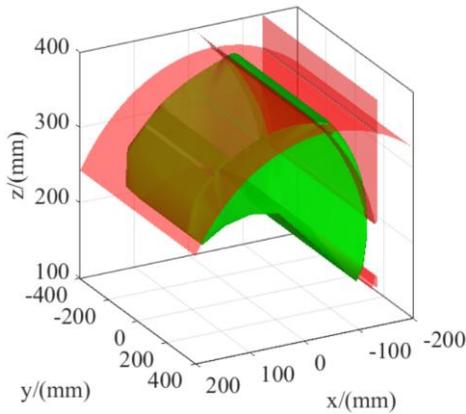
(a) Serial singularity surface in workspace

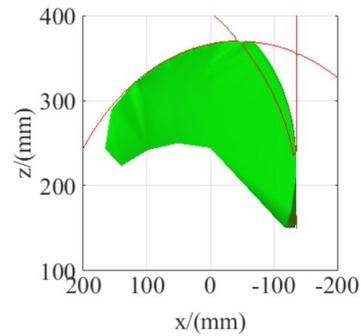
（b）Projection view of the *xoz* section of serial singularity

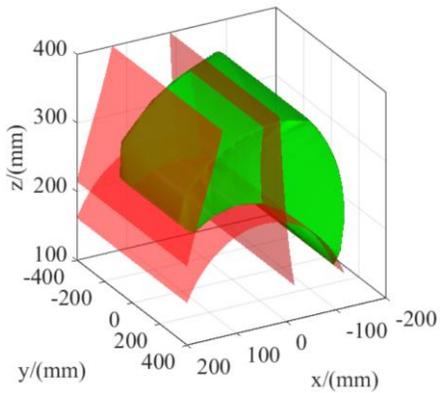
（c）Parallel singularity surface in workspace

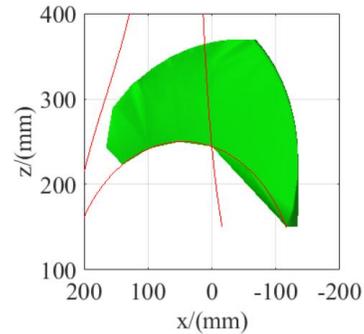
（d）Projection view of the *xoz* section of parallel singularity

Figure 11. Singularity surfaces in the workspace

It is noted that Fig.8 shows the singularity surfaces that contains all the inverse kinematic solutions inside the workspace, while Fig.11 shows the positional relationship between the useful workspace and the singularity surface within the range of the useful workspace. In Fig.11, the green solid means the useful workspace, and the red surface means the singularity surface inside the workspace.

## 5. Conclusions

In this paper, a novel three-translation parallel manipulator is presented, which combines the advantages below and potentially outperforms in application compared to the TPM counterparts.

① The proposed TPM can have fewer joints and components than the well-known Delta robot, which reduces the structural complexity. Compared to the 3UPU or 3RRC counterparts, the asymmetrical architecture of the proposed TPM allows partially decoupled motion to ease motion control and trajectory planning.

②The lengths of guide rails on which the actuated-joints are located can have finitely large lengths as possible according to the application requirements, which determines a large workspace rather than its fully symmetric counterparts.

③The TPM has symbolic solutions to the direct and inverse kinematic problems. The symbolic direct position solution is beneficial for error analysis, workspace analysis, and velocity/acceleration and dynamics analysis.

According to the kinematics modelling principle based on topological characteristics, the virtual variable α can be solved by the geometric constraints and the topological constraints of the first loop with a positive constraint degree, instead of the constraints of the second loop with a negative constraint degree. This is a key factor for the TPM in this paper to be able to obtain the symbolic solution, and it is also an advantage of the TPM's topology.

The Jacobian matrix is derived to identify the singular configurations, and the singular loci are found and visualized by means of a Gröbner-based elimination operation, to show the singularity-free workspace. From the two aspects of theoretical calculation and Adams simulation, the motion characteristics of the TPM are analyzed, which provides the fundamentals for the stiffness, dynamics and prototyping of the proposed TPM.

**ACKNOWLEDGMENTS:**

The reported work is supported by the NSFC No.51975062.